\documentclass{article}

\usepackage{natbib}
    \usepackage[final]{neurips_2023_arxiv}

\usepackage[utf8]{inputenc} 
\usepackage[T1]{fontenc}    
\usepackage{hyperref}       
\usepackage{url}            
\usepackage{booktabs}       
\usepackage{amsfonts}       
\usepackage{nicefrac}       
\usepackage{microtype}      
\usepackage{xcolor}         
\usepackage{graphicx}
\usepackage{amsmath,amsthm,amssymb}

\DeclareMathOperator*{\argmin}{argmin}
\DeclareMathOperator*{\expectation}{\mathbb{E}}

\newcommand{\R}{\mathbb{R}}  
\newcommand{\Z}{\boldsymbol{Z}}

\newcommand{\Q}{\boldsymbol{Q}}

\newcommand{\X}{\boldsymbol{X}}
\newcommand{\Y}{\boldsymbol{y}}
\newcommand{\U}{\boldsymbol{U}}
\newcommand{\W}{\boldsymbol{w}}
\newcommand{\Laplacian}{\boldsymbol{L}}
\newcommand{\normLaplacian}{\boldsymbol{\bar{\Laplacian}}}
\newcommand{\Lambdab}{\boldsymbol{\Lambda}}
\newcommand{\graph}{\mathcal{G}}

\newcommand{\tran}[2]{q_{#1#2}}  
\newcommand{\stat}[1]{\pi_{#1}}  
\newcommand{\joint}[2]{\rho_{#1#2}}  
\newcommand{\adj}{\boldsymbol{A}}
\newcommand{\normadj}{\boldsymbol{\bar{\adj}}}

\newcommand{\degree}{\boldsymbol{D}}
\newcommand{\obs}[1]{x_{#1}}
\newcommand{\I}{\boldsymbol{I}}

\title{Spectral Temporal Contrastive Learning}

\author{%
  Sacha Morin$^{\dagger, 1,2}$, Somjit Nath$^{\dagger, 2,3}$, Samira Ebrahimi Kahou$^{2,3,4}$, Guy Wolf$^{1,2,4}$\\
  $^1$Université de Montréal, $^2$Mila, $^3$École de technologie supérieure, $^4$CIFAR AI Chair \\
}

\begin{document}

\maketitle

\def\thefootnote{$\dagger$}\footnotetext{Equal Contribution}\def\thefootnote{\arabic{footnote}}
\begin{abstract}
Learning useful data representations without requiring labels is a cornerstone of modern deep learning. Self-supervised learning methods, particularly contrastive learning (CL), have proven successful by leveraging data augmentations to define positive pairs. This success has prompted a number of theoretical studies to better understand CL and investigate theoretical bounds for downstream linear probing tasks. This work is concerned with the temporal contrastive learning (TCL) setting where the sequential structure of the data is used instead to define positive pairs, which is more commonly used in RL and robotics contexts. In this paper, we adapt recent work on Spectral CL to formulate Spectral Temporal Contrastive Learning (STCL). We discuss a population loss based on a state graph derived from a time-homogeneous reversible Markov chain with uniform stationary distribution. The STCL loss enables to connect the linear probing performance to the spectral properties of the graph, and can be estimated by considering previously observed data sequences as an ensemble of MCMC chains.
\end{abstract}

\section{Introduction}
In recent years, Self-Supervised Learning (SSL) has gained popularity in fields with a wealth of unlabeled data such as computer vision and natural language processing. 
SSL has proven empirically successful: linear probing of ImageNet representations learned by both contrastive and non-contrastive SSL methods has shown surprising performance, being almost on par with supervised models \citep{bardes2021vicreg}. 
This success has prompted some authors to investigate SSL objectives from a theoretical perspective. Of particular interest to us is a recent work by \citep{haochen2021} which formulates a learning objective based on a data augmentation graph and applies spectral graph theory tools to study optimal representations and derive error bounds for linear probing.

While data augmentation is perhaps the most common way of defining positive and negative pairs in a contrastive setting, potentially any known structure over the data points can be leveraged for contrastive learning. In particular, temporal contrastive learning (TCL) relies on the temporal structure of the data: data points that are close in time in observed sequences are treated as positives. Representation learning is usually achieved over a set of previously observed sequences. This paradigm finds application for representation learning for videos \citep{dave2022tclr}, reinforcement learning (RL) \citep{erraqabi22a}, navigation \citep{yang2020plan2vec, morin2023one}, and potentially any other setting where sequential data is available. Observe that we are interested in \textbf{learning an embedding for each observation}, as opposed to learning a single embedding for the entire sequence of observations. This work aims to adapt the theoretical findings of \cite{haochen2021} to the temporal setting. Our main contributions are as follows:

\begin{enumerate}
    \item We derive a TCL objective, which we dub Spectral Temporal Contrastive Learning (STCL);
    \item STCL is based on a state graph which is itself derived from a time-homogenous reversible Markov Chain;
    \item Under a uniform stationary distribution assumption, the STCL minimizers can be directly characterized in terms of the eigenvectors of the state graph Laplacian, which grants access to additional tools to bound the linear probe error.
\end{enumerate}

\section{Background}
\paragraph{Self-Supervised Learning.} Given an input data matrix $\X$, \textbf{self-supervised learning} (SSL) can be summarized as learning a ``useful" data representation $\Z$ in the absence of clearly defined labels, often using a deep network. In practice, usefulness is subsequently measured by using $\Z$ in some supervised downstream task to predict some labels $\Y$ using a linear model. This evaluation procedure is known as \textbf{linear probing} and ensures that relevant information (i.e., $\Y$) can be recovered from $\Z$ using a low capacity model. Naturally, $\Y$ cannot be used while fitting $\Z$.

\paragraph{Theoretical Works on Contrastive Learning.} \cite{arora2019} provides guarantees on a linear classification downstream task for a representation, $\Z$ learned by contrastive learning. One of the important assumptions of this work is that the positive pairs have the same conditional distribution, which basically translates to the positive pairs of augmentations being conditionally independent given the class label of the downstream task. There are some additional works~\cite{lee2020} in this aspect that focus solely on reconstruction-based methods, where learning $\Z$ happens by reconstructing the current input $\X$ in a lower dimensional space. In such methods, the authors assume conditional independence with respect to latent variables, for example, when we want to classify scenery like forests, seas, or deserts from images, we can use the background color as a latent variable. 
More interestingly, they weaken this assumption and provide bounds for approximate conditional independence(~\cite{lee2020}, Theorem 4.2). Recently,~\cite{tosh2021} finds guarantees for linear probing by assuming there exists a hidden variable such that the positive pairs are conditionally independent given that hidden variable.

Of particular interest to our paper is the recent trend of analyzing contrastive learning through the lens of spectral graph theory \citep{haochen2021, balestriero2022contrastive}. A natural way of manipulating positive relationships over a dataset is to treat them as edges in a graph. Let $\graph$ be a graph with the rows of $\X$ (data points) as vertices and where edges represent a positive relationship. 

\cite{haochen2021} leverages the graph framework to propose a Spectral Contrastive Learning objective with provable guarantees for a downstream classification task. They first introduce the notion of \textbf{population augmentation graph} where $\X$ in fact consists of all possible data points and their augmented views \footnote{They consider an exponentially large, but finite data space, e.g., all vectors in a subset of $\mathbb{R}^d$ with finite precision representations.}. They then formalize the following intuition: it is substantially easier to go from one image to the other using augmentations if both images belong to the same class according to the downstream labels $\Y$. This implies that the different subgraphs in $\graph$ corresponding to the different classes in $\Y$ will have strong intra-connectivity but weak inter-connectivity, which in turn ensures that some eigenvectors of the Laplacian of $\graph$ will align with the class structure \citep{ng2001spectral}. 

The authors then introduce their SSL objective and show that any minimizer $\Z^*$ will correspond to the bottom eigenvectors of the Laplacian of $\graph$ , up to a left scaling transform and a right linear transform. They then derive a bound on the linear probe error in the population setting that essentially depends on how well the classes $\Y$ turn out to be clustered in $\graph$ (Theorem 3.8). They further consider additional finite-sample generalization bounds.


\paragraph{The Graph Laplacian in RL.} A number of papers formulate a state graph similar to ours and rely on some elements of spectral graph theory to design learning objectives in the context of RL \cite{machado2017laplacian, erraqabi22a}. LapRep~\citep{wulap} is particularly relevant to our work, as it formulates a similar state graph based on a Markov chain and derives a contrastive learning objective using the classical \textit{graph drawing} objective of \citep{yehudaspec} as a starting point, which should also be minimized by the Laplacian eigenvectors. Our final objective differs substantially from the LapRep one as we rely on the matrix factorization perspective of \citep{haochen2021} to derive \textbf{both} the positive and the negative terms in the contrastive loss, while the LapRep negative term is obtained by relaxing the original orthogonality constraint of graph drawing. Nevertheless, the LapRep method would be an interesting benchmark in future empirical comparisons of STCL.

\section{Spectral Temporal Contrastive Learning}
This section will adapt the theory of \cite{haochen2021} to the temporal setting and derive the Spectral Temporal Contrastive Loss (STCL). We will begin by defining a Markov chain and an associated state graph $\graph$ in Subsection \ref{subsec:graph}. Using this state graph as a temporal analog of the data augmentation graph, we will introduce the population matrix factorization loss from \cite{haochen2021} in Subsection \ref{subsec:pop} and characterize its minimizers in terms of the eigenvectors of the normalized graph Laplacian. Finally, Subsection \ref{subsec:empiricalLoss} shows how a contrastive loss can estimate the population loss without observing $\graph$ using samples from the Markov Chain.

\subsection{Markov Chain \& State Graph Definitions}
\label{subsec:graph}

Consider a time-homogeneous Markov chain over discrete finite states $S=\{i\}_{i=1}^N$ with transition probabilities $\tran{i}{j}$ and some associated observations  $\obs{i} \in \mathcal{X}$. We further assume a unique stationary distribution $\stat{}$ and let $\joint{i}{j}:=\stat{i}\tran{i}{j}$. We will now build a weighted state graph $\graph$ under the assumption that the chain is reversible: $\joint{i}{j}=\joint{j}{i}$. Using $S$ as vertices we define an adjacency matrix $\adj$ such that $\adj_{ij} = \joint{i}{j}.$
The reversibility property ensures that $\adj$ is symmetric. Moreover, if the first state in the chain is distributed according to $\stat{}$, then $\joint{i}{j}$ can also be interpreted as a joint distribution over transitions, with $\stat{}$ acting as a marginal over states and $\tran{}{}$ as a conditional. Computing the elements of the diagonal degree matrix $\degree$, therefore, corresponds to marginalizing states, i.e.
\begin{align}
    \degree_{ii} = \sum_{j}\joint{i}{j}=\stat{i} \implies \degree = \text{diag}(\pi).
\end{align}
\vspace{-1.5em}

Equipped with the matrix $\degree$, we can define the normalized adjacency matrix as
$
    \normadj = \degree^{-\frac{1}{2}}\adj \degree^{-\frac{1}{2}} 
$
and the normalized Laplacian as
$
    \normLaplacian = \I - \normadj.
$ Observe that $\normadj$ and $\normLaplacian$ share the same eigenvectors, and if $\lambda$ is an eigenvalue of $\normadj$ then $1 - \lambda$ is an eigenvalue of $\normLaplacian$. This fact will allow to interchangeably use the top eigenvectors of $\normadj$ or the bottom eigenvectors $\normLaplacian$ in theoretical arguments.

\subsection{STCL in the Population Setting}
\label{subsec:pop}
\paragraph{Population Objective.}
We consider the problem of learning vertex Euclidean representations from the high dimensional observations $\X$  using a deep encoder $f:\mathcal{X}\to \R^k$. In this section, we assume knowledge of $\graph$ and $\normadj$. We further assume in the population setting that we have exactly one observation per state, which allows to build 
 a population representation matrix $\Z \in \R^{N\times k}$ by stacking $f(\obs{i})$ as rows. To learn our feature extractor $f$, we will adapt the objective from \cite{haochen2021} to our Markov chain setting. Consider the low-rank matrix factorization objective
\begin{align}
\label{eq:mx_loss}
    \mathcal{L}_{mf}(\Z) = \| \normadj - \degree^{\frac{1}{2}} \Z \Z^\top \degree^{\frac{1}{2}\top} \|_F^2.
\end{align}
and the minimizers $\Z^* = \argmin_{\Z \in \R^{N \times k}} \mathcal{L}_{mf}(\Z)$.
The rescaling by $\degree^{\frac{1}{2}}$ effectively means that we consider rescaled representations $\sqrt{\stat{i}}f(\obs{i})$ in the learning objective. As we will see, this rescaling will prove critical in the derivation of a sampling-based learning objective (Section \ref{subsec:empiricalLoss}), but will complicate the analysis of $\Z^*$.

The matrix factorization approach allows us to characterize the minimizers $\Z^*$ in terms of the eigenvectors of $\normadj$, which in turn provides rich information on the geometry of $\graph$. Indeed, 
the Eckart–Young–Mirsky Theorem~\citep{eckart1936approximation} ensures that any rank $k$ minimizer $\Z^*$ of Equation \ref{eq:mx_loss} can be expressed as $\Z^*=\degree^{-\frac{1}{2}}\U_k \Lambdab_k^{\frac{1}{2}}\Q$ where, $\U_k \in \R^{N \times k}$ denotes the top $k$ eigenvectors of $\normadj$, $\Lambdab \in \R^{k \times k}$ is the truncated diagonal matrix of the top eigenvalues, and $\Q \in \R^{k \times k}$ is an orthogonal matrix. $\U_k$  can be equivalently understood as the bottom eigenvectors of $\normLaplacian$.


\paragraph{Linear Probing.} Following the SSL setting, we then freeze the representations $\Z$ and perform linear probing by minimizing a regression loss with respect to some target vector $\Y \in \R^{N}$:
\begin{align*}
   \mathcal{C}(\W) &= \min_{\W \in \R^{k}} \| \Y - \Z^* \W\|^2 \\
   &= \min_{\W \in \R^{k}} \| \Y - \degree^{-\frac{1}{2}}\U_k \W\|^2 
\end{align*}
where the second line results from the fact that both $\Lambdab_k^{\frac{1}{2}}$ and $\Q$ are invertible, which allows to freely reparametrize $w$. Linear probing can in principle reach zero error whenever $\Y \in span(\degree^{-\frac{1}{2}}\U_k)$.

The matrix $\degree^{-\frac{1}{2}}$ prevents a direct connection to the spectral properties of $\graph$, since the normalized adjacency matrix eigenvectors $\U_k$ do not directly span the space of interest. In \cite{haochen2021}, the authors overcome this limitation by considering a classification task. The weights in $\degree^{-\frac{1}{2}}$ are positive and only act as temperature parameters in the softmax used by the classifier probe, with no impact on the logit rankings. For the purpose of training a linear classifier, the authors claim that the representation $\degree^{-\frac{1}{2}}\U_k$ is as useful as $\U_k$.

\paragraph{Uniform Stationary Distribution.} In the regression setting, $\degree^{-\frac{1}{2}}$ matters more. Observe that in the specific case where $\stat{}$ is uniform over $S$, $\degree^{-\frac{1}{2}}$ effectively acts as a scalar scaling of all dimensions by $\frac{1}{\sqrt{N}}$. In this case, $span(\U_k)$ properly describes the set of linear tasks achievable with the learned representation and allows to directly apply spectral graph theory. We further observe that under the uniform $\stat{}$ assumption, $\normadj = \frac{1}{N} \adj$ and the normalization step does not affect the eigenvectors of the original adjacency and Laplacian matrices.

\begin{figure}[ht]
\centering
\begin{minipage}[t]{0.4\textwidth}
\centering
\includegraphics[width=0.65\textwidth]{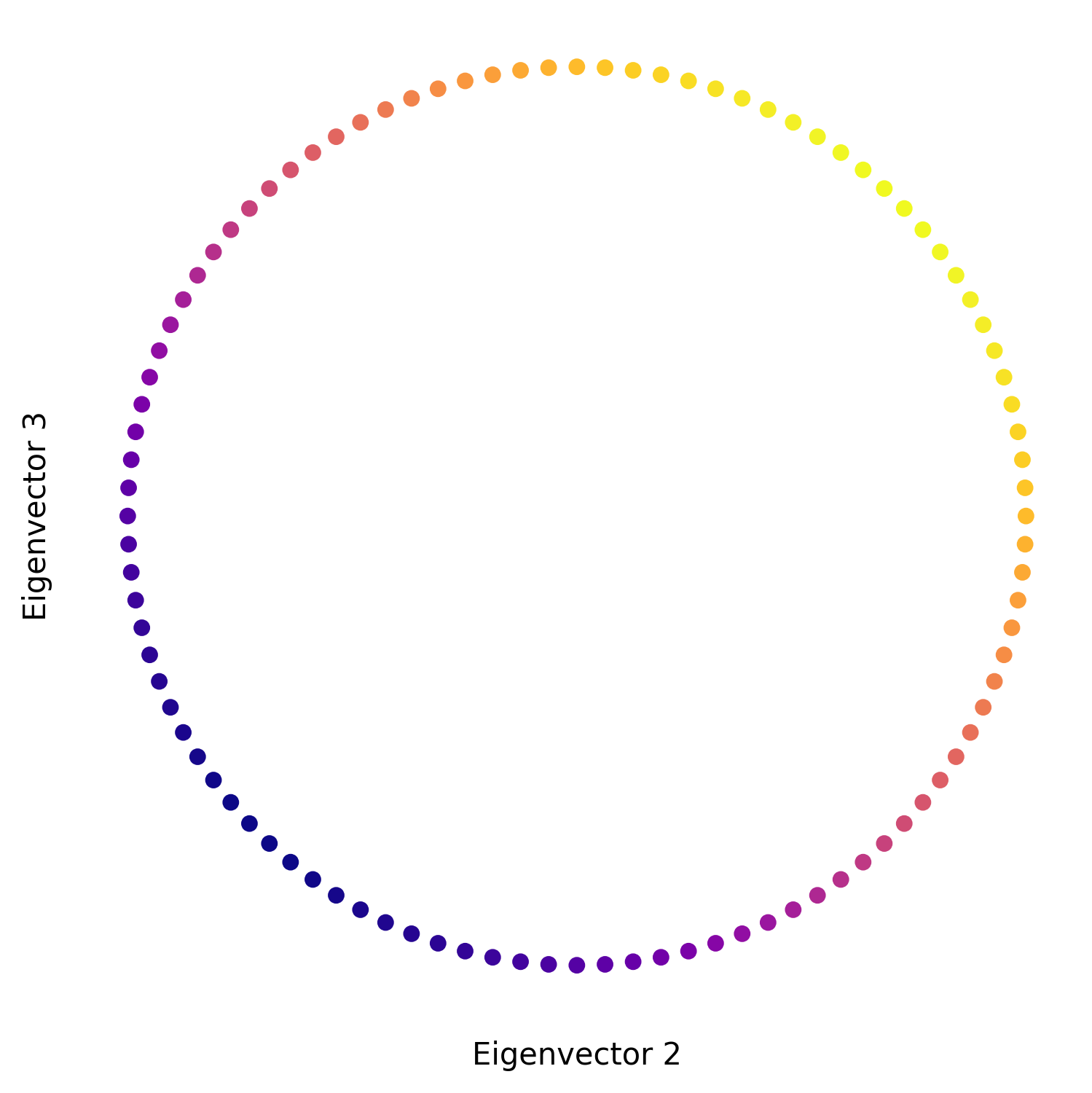}
\caption{Top Eigenvectors of $\normadj$ for a Markov Chain over a \textbf{ring graph} with 100 states.}
\label{fig:ring}
\end{minipage}
\hspace{2em}
\begin{minipage}[t]{0.4\textwidth}
\centering
\includegraphics[width=0.65\textwidth]{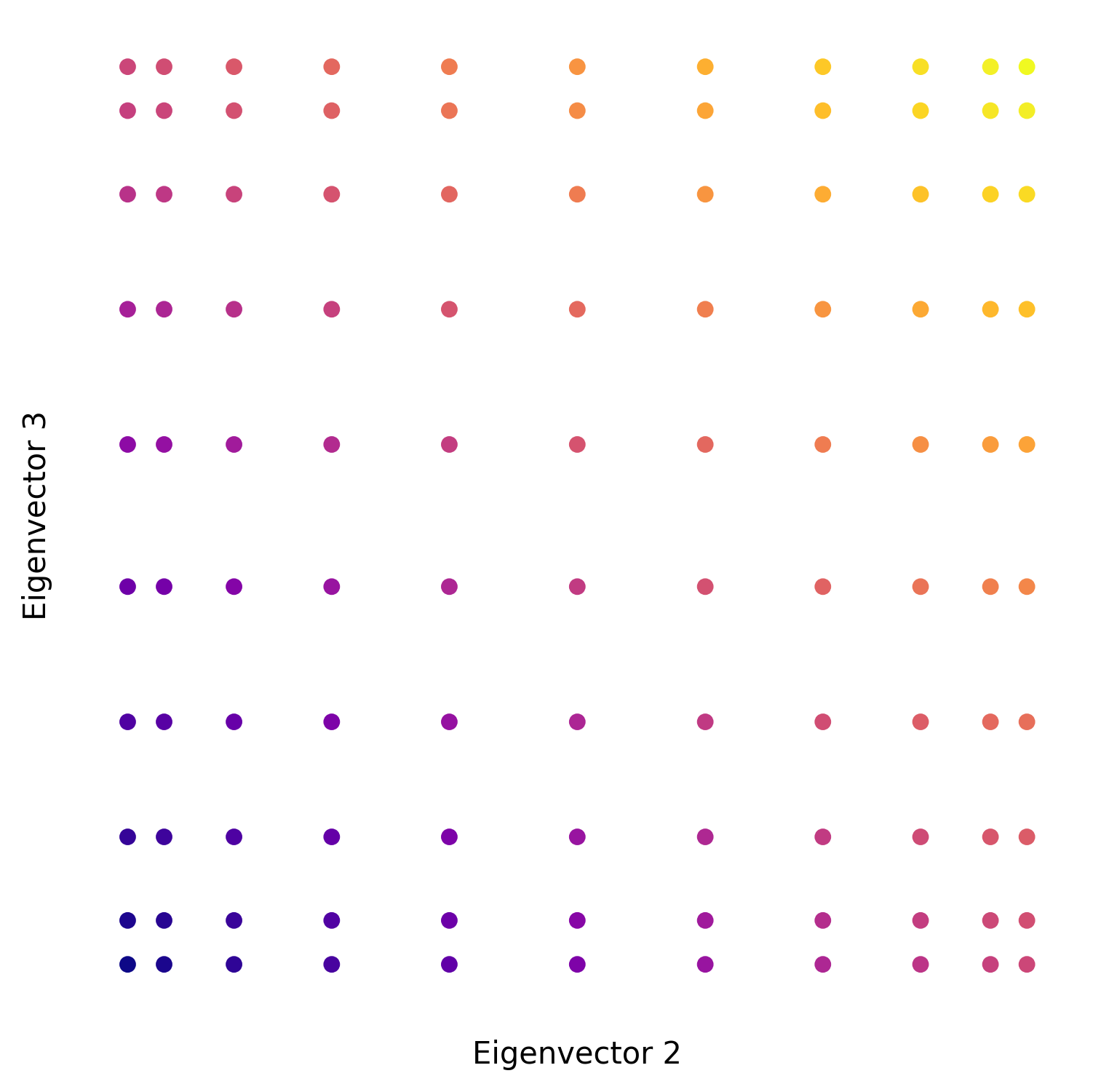}
\caption{Top Eigenvectors of $\normadj$ for a Markov Chain over a 11x10 \textbf{grid graph}.}
\label{fig:grid}
\end{minipage}
\end{figure}

\paragraph{Tasks.} Describing the set of ideal linear tasks as $\Y \in span(\U_k)$ is an appealing proposition since $\U_k$ is known to provide a Euclidean embedding that captures the geometry of $\graph$ \citep{spielman2019spectral}. To illustrate this point, we show the top eigenvectors of $\normadj$ for a ring graph in Figure~\ref{fig:ring} and a 2D grid graph in Figure~\ref{fig:grid}. Pairs of state with high transition probabilities are mapped closely in terms of Euclidean distance. As we will further explore in our experiments in Appendix~\ref{subsec:experiments}, this property can be leveraged for state/pose prediction tasks that strongly correlate with the underlying state graph connectivity. $\U_k$ is also known to encode the cluster structure of the graph when such clusters exist \citep{ng2001spectral}. In navigation tasks, this fact can be exploited to predict different "rooms" in the environment and learn RL options \citep{machado2017laplacian}.



\subsection{STCL Loss Estimator}
\label{subsec:empiricalLoss}
\paragraph{Contrastive loss.}
In practice, we do not have access to the graph $\graph$ or all the states in $S$ to train $f$. However, it turns out that $\mathcal{L}_{mf}$ (Equation \ref{eq:mx_loss}) can be expanded as 
\begin{align*}
    \mathcal{L}_{mf}(\Z)
    & \propto  - 2\sum_{ij} \joint{i}{j}f(x_i)^\top f(x_j) + \sum_{ij} \stat{i}\stat{j}(f(x_i)^\top f(x_j)^2 \stepcounter{equation} \tag{\theequation}\label{eq:loss} \\
    \mathcal{L}(f) &:= -2\expectation_{(i, j) \sim \joint{}{}}[f(x_i)^\top f(x_j)] + \expectation_{i \sim \stat{}, \; j \sim \pi} [(f(x_i)^\top f(x_j))^2].  \stepcounter{equation}\tag{\theequation}\label{eq:expectedloss}
\end{align*}
We can see a contrastive loss emerging from Equation \ref{eq:expectedloss}: the first term pushes positive pairs sampled from the transition joint $\joint{}{}$ to be aligned. The second one samples ``negative pairs" independently from the stationary distribution $\pi$ and is minimized whenever their representations are orthogonal.

\paragraph{Sampling.} We cannot directly sample from $\stat{}$ and $\joint{}{}$ to estimate $\mathcal{L}$. However, it is common to assume in TCL settings that we have access to a previously collected set of sequences $\{(x^m_{t})_{t=1}^T\}_{m=1}^M$ which we can use to train $f$. Examples of this setting includes exploring an environment with a robot~\citep{morin2023one} or pretraining on human videos for offline RL~\citep{ma2022vip}.

We consider the observed sequences as chains sampled from the Markov chain defined in Section~\ref{subsec:graph}. MCMC theory suggests that a very long chain or an ensemble of chains, ideally from different starting states, should eventually mix and allow to approximately sample from $\stat{}$ (and from $\joint{}{}$ by first sampling $x_t$ from $\stat{}$ and then sampling $x_{t+1}$ from observed transitions from $x_t$). For our experiments, we simply treat our observations as samples from $\stat{}$, even if mixing is not guaranteed. We provide early experimental results in Section \ref{subsec:experiments}.

\section{Conclusion and Future Works}
In this work, we extended the results of \cite{haochen2021} to the temporal contrastive learning setting by considering the state graph induced by an underlying Markov chain.  In future work, we hope to leverage Lemma B.6 of \cite{haochen2021} ---and potentially additional assumptions on the task $\Y$ and the graph $\graph$--- to obtain more explicit and interpretable guarantees. Moreover, the uniform stationary assumption is somewhat restrictive. We hope that a more general argument can handle the left scaling by $\degree^{-\frac{1}{2}}$ to better characterize the span of the STCL minimizers. We also aim to relax the reversibility assumption and extend the analysis to continuous state spaces.

\bibliography{ref}

\section{Appendix}

\subsection{Experiments}
\label{subsec:experiments}
\begin{figure}[ht]
\centering
\begin{minipage}[t]{0.9\textwidth}
\centering
\includegraphics[width=1.0\textwidth, height=0.5\textwidth]{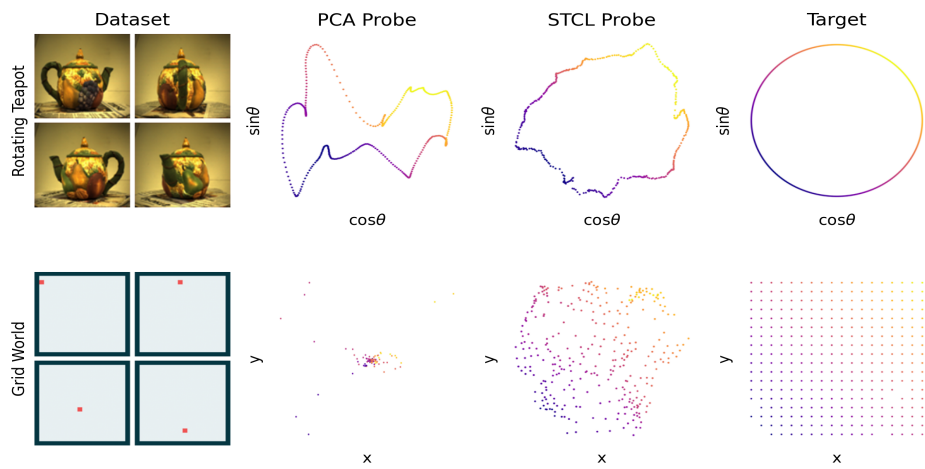}
 \caption{Outputs of the linear probe for the PCA baseline (second column) and STCL (third column). Ground truth targets are shown in the last column.}
 \label{fig:allplots}
\end{minipage}
\end{figure}

We consider 2 toy experiments. The first dataset consists of images of a 3D Teapot model rotated over 400 different angles \cite{weinberger2004learning}, and the downstream task is to predict the Teapot angle pose $(\cos\theta, \; \sin\theta)$. The second dataset is based on images generated by a Grid World environment, where the probe task will be to predict the $(x,\; y)$ coordinates of the agent. We collect random trajectories with random uniform policies in both environments and train STCL with a Resnet-18 backbone and representations of dimension $k=8$. As a baseline, we use PCA to extract the top 8 principal components and run the same linear probe. We visualize the outputs of the probes in Figure~\ref{fig:allplots} and their $R^2$ metrics in Table~\ref{table:table_jackal}.

\begin{table}[!hptb]
\medskip
\centering
\caption{Coefficient of determination ($R^2$) of a supervised linear probe for PCA, Spectral Temporal Contrastive learning (STCL) and Ground Truth (GT). Please refer to main text for details.
}
\setlength{\tabcolsep}{5.5pt}
\begin{tabular}{lccc}
\toprule
&  PCA Probe $\uparrow$ & STCL Probe $\uparrow$ & GT $\uparrow$  \\
\midrule
Teapot & 0.80 & 0.98 & 1.00\\
Grid World & 0.19 & 0.96 & 1.00\\
 \bottomrule
\end{tabular}
\label{table:table_jackal}
\end{table}

\end{document}